**Improving Autonomous Vehicle Mapping and Navigation in Work Zones Using Crowdsourcing Vehicle Trajectories**

**Hanlin Chen, ORCID: 0000-0001-6508-7715**
Lyles School of Civil Engineering, Purdue University
550 Stadium Mall Drive, West Lafayette, IN 47907
Email: chen1368@purdue.edu

**Renyuan Luo**
Krannert School of Management, Purdue University
403 W State St, West Lafayette, IN, 47906
Email: luo266@purdue.edu

**Yiheng Feng, ORCID: 0000-0001-5656-3222**
Lyles School of Civil Engineering, Purdue University
550 Stadium Mall Drive, West Lafayette, IN 47907
Email: feng333@purdue.edu

Word Count: 5369 words = 5369ss words

Submitted on August 1st, 2022

**ABSTRACT**
Prevalent solutions for Connected and Autonomous vehicle (CAV) mapping include high definition map (HD map) or real-time Simultaneous Localization and Mapping (SLAM). Both methods only rely on vehicle itself (onboard sensors or embedded maps) and can not adapt well to temporarily changed drivable areas such as work zones. Navigating CAVs in such areas heavily relies on how the vehicle defines drivable areas based on perception information. Difficulties in improving perception accuracy and ensuring the correct interpretation of perception results are challenging to the vehicle in these situations. This paper presents a prototype that introduces crowdsourcing trajectories information into the mapping process to enhance CAV's understanding on the drivable area and traffic rules. A Gaussian Mixture Model (GMM) is applied to construct the temporarily changed drivable area and occupancy grid map (OGM) based on crowdsourcing trajectories. The proposed method is compared with SLAM without any human driving information. Our method has adapted well with the downstream path planning and vehicle control module, and the CAV did not violate driving rule, which a pure SLAM method did not achieve.

**INTRODUCTION**
Temporarily changed drivable areas are a major hurdle for large-scale deployment of CAVs, as drivable area inference on complex cases remains a challenge. Among them, the work zone is one of the typical cases. Work zones are common in all types of roadways in the United States (*1*). They serve the function of "construction, maintenance or utility work" according to the definition of Federal Highway Administration (FHWA) (*2*). Many work zones do not completely block the entire traffic flow while executing the maintenance or upgrading work. This leads to temporary changes in the drivable area and requires the adaptation of autonomous vehicles when driving through. As the demand capacity ratio is usually increased in work zones due to reduced capacity, effective navigation through the work zones is crucial for CAVs to avoid being moving traffic bottlenecks while also ensuring the safety of other road users (*3*). Effective navigation of CAVs requires accurate mapping, a challenging task in work zones.

Mapping serves an important function in autonomous driving systems. At the microscopic level, it provides additional information over real-time perception (*4*). This enables CAVs to perform self-localization and guides the planning module to generate a feasible reference trajectory for the near future. Common tasks for roadway detection from the perception system include lane detection, instance segmentation to extract drivable areas, and so on. For typical perception scenarios, the drivable area is defined by lane markings. In some edge cases, however, the lane markings are not visible or overlapping, which cannot be used to define the drivable areas. In these cases, other markers are placed on the roadway to define drivable areas (e.g., traffic cones in work zones). With additional markers, drivable area inference is usually easy for human drivers but may not be easy from a computer vision's perspective.

The challenge for navigating CAVs within the work zone is mainly about the drivable area identification given road geometry and perception information. If mapping and localization rely on HD maps, problems arise when the original HD map can't be utilized since the drivable area is temporarily changed. Updating HD maps cannot be conducted in real-time and is expensive. Another option for autonomous vehicles is the real-time inference of the drivable area through the perception system. However, real-time mapping technologies such as Simultaneous Localization and Mapping (SLAM) also have limitations. For SLAM, common problems include but are not limited to divergence given biased estimation, presence of dynamic obstacles, and drifting in localization (*5*). Specifically for traffic cone detection, the accuracy ranged from 58.5%-75.8% for 2D image object detection (*6*). More significant problems arise with de-shaped, tilted, or missing traffic cones within the work zones, let alone different weather and lighting conditions (*6*). Even if the traffic cone detection can achieve 100% accuracy without prior information, CAVs may have difficulty identifying drivable areas.

An example is shown in Figure 1. Without prior information, drivers may think the right most lane and the middle lane are both drivable, while in reality, only the middle lane is allowed. In this case, even human drivers may have difficulties in identifying correct lane(s) to drive without explicit instructions. Pure perception and single vehicle-based solutions may have bigger challenges.

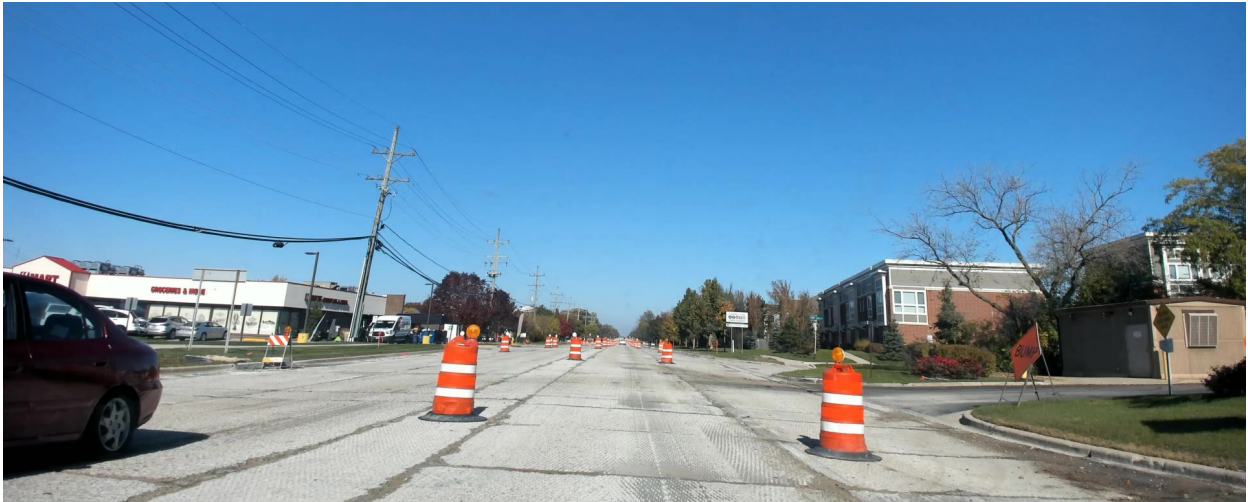

**Figure 1 Work zone with confusing drivable areas.**

Crowdsourcing trajectories can be a new source of information for drivable area inference when it comes to inferencing drivable areas. Crowdsourcing trajectories can be obtained from various sources such as connected vehicles (CVs), ride-sharing companies (e.g., Lyft, Uber), and data companies (e.g., INRAX, Wejo). They can "teach" CAVs where are the drivable areas in the work zone and how to navigate safely through human driver's demonstrations. Due to the lack of explicit rules in many work zone driving scenarios (e.g., where and when to merge to another lane), driving patterns in these areas are more from a group voting perspective. These uncertainties pose more challenges for CAVs. Moreover, crowdsourcing trajectories can reduce inference errors from pure perception and adapt to changes in drivable areas that had not been included in the training data.

This paper presents a prototype of introducing crowdsourcing trajectories into the mapping process to improve CAV navigation in work zones. To the best of our knowledge, there had not been prior work that integrated crowdsourcing trajectories into the mapping process. A Gaussian Mixture Model (GMM) is applied to represent the statistical distributions of crowdsourcing trajectories. The drivable area is generated from sampling trajectory points from the fitted GMM. To illustrate the advantage of our work, we construct a similar drivable area generation pipeline following the algorithms implemented in Autoware.AI, a standard middleware system for autonomous driving. We compare the two methods in the CARLA simulation and show the advantages of incorporating crowdsourcing information into the map generation process.

The remainder of the paper is described as follows. Section 2 briefly introduces the literature related to our work. Section 3 elaborates on the methodology we used in this work. In section 4, we conduct a case study comparing the proposed crowdsourcing trajectory-based drivable area generation method with the standard SLAM-based method in a work zone in the CARLA simulation platform. Discussion and future work are provided in section 5, with conclusions in section 6.

**LITERATURE REVIEW**
Perception, localization, and mapping are critical components for CAVs to know their location and surrounding environment. The common practice in CAV mapping relies on either a prior HD map or SLAM(*4*). Considering the temporarily changed drivable area, we care most about the geometric and semantic information for the area of interest (*4*). The identification of drivable areas is essential for the path planning module. SLAM has better adaptability over HD maps as SLAM performs real-time mapping of the surrounding environment. Common SLAM techniques for autonomous vehicles utilize point cloud data and the precise location of the ego vehicle (e.g., obtained from RTK GPS). Such techniques can be categorized as point-based registration like point cloud matching based on normal distribution transform (NDT matching)(*7*) and feature-based point cloud registration like Lidar Odometry and Mapping in Real-time (LOAM)(*8*).

From the perception's perspective, current work aiming at temporary changes in drivable areas mainly targeted the identification of traffic cones in work zones. These identification works include image data (*9*)(*10*) and point cloud data (*11*). There had been some perception-related work using roadside sensors or UAS (*12*)(*13*) , yet these works focus on the safety management within work zones, not adaptive control for CAVs. There are some works targeting detecting irregular-shaped drivable areas given camera data. These works used the color-based method (*14*) or the lane-related detection method (*15*)(*16*). Although some open datasets exist, including the work zone scenarios in the training set (*17*) (*18*), the challenge remains even after all the traffic cones within a work zone can be successfully detected. Reasoning a drivable area given a set of traffic cones seems easy for a human but is still a challenge for computer vision techniques, as shown in the example in Figure 1.

Some attempts address the path planning problem within temporarily changed drivable areas. However, they either rely on accurate lane detection (*19*) or accurate detection of dynamic and static objects (*20*). To the best of our knowledge, there is no prior work targeting the reasoning process given detected traffic cones and output a predicted drivable area, nor integrating human driving information into the mapping process. There are some studies utilizing existing trajectories to predict possible future trajectories *(21) (22)*, and we borrow the idea from these kinds of works. Wiest (*23*) built a GMM given multiple historical trajectories, and the prediction of future trajectory is based on the fitted GMM. Another work utilizing GMM on trajectory prediction is done by Luo (*24*). Their work initialized several GMMs, each describing a cluster of human motion trajectories. For the online learning framework, newly recorded trajectories are either used to update a GMM's parameter or to initialize a new GMM if none of the existing GMM models can describe the new trajectory. Both works utilized GMM to describe common features of the vehicle trajectories.

## Methodology

### System Overview
Figure 2 shows the components of our work. The modules colored in blue indicate the standard practice for CAVs. We adopt the current practice from Autoware.AI, which is mapping and trajectory generation from SLAM. The modules colored in orange indicate our work. As shown in Figure 2, we utilize information from crowdsourced trajectories and generate drivable areas which can be utilized by standard path planners for CAVs.

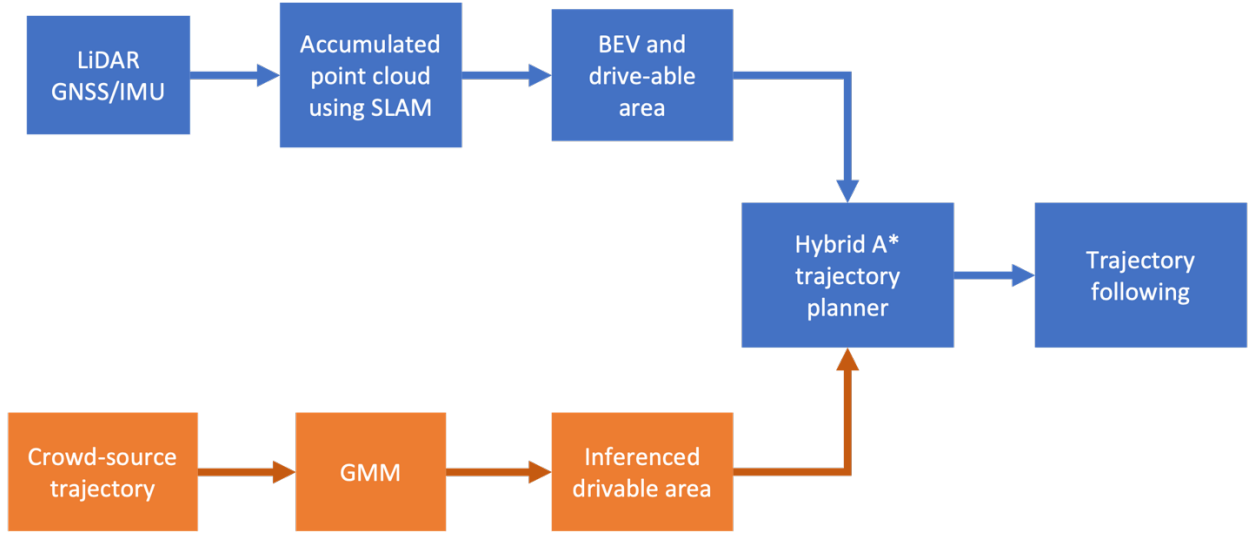

**Figure  System components in our work**

In the standard practice, we apply Simple Morphological Filter (SMRF) (*25*) and Lidar Odometry and Mapping in Real-time (LOAM) (*8*) in the standardized SLAM. Based on the accumulated point cloud map, we obtained a BEV-based drivable area from SLAM. In comparison, we obtained a GMM model from crowdsourced trajectories and inference the drivable area in the form of an occupancy grid map (OGM). Based on the OGM obtained either from SLAM or from our work, we generate a reference trajectory using the Hybrid A* planner (*26*)(*27*). Finally, the pure pursue algorithm is applied as the trajectory following model. We will discuss the methods used in detail in the following sections.

**Gaussian Mixture Model (GMM)**

Denote $\boldsymbol{x}$ is a two-dimensional Gaussian random variable representing vehicle trajectory points in the 2D space. A Gaussian Mixture Model (GMM) is a probabilistic model that consists of several Gaussian components in the form of **Equation 1**, Where $\boldsymbol{\mu}_k$ and $\boldsymbol{\Sigma}_k$ are the mean and covariance matrix of the $k^{th}$ component, and $\pi_k$ is the proportion of the $k^{th}$ component with $\pi_k \geq 0$ $and$ $\sum \pi_k = 1$.

$$p(\boldsymbol{x}) = \sum_{k=1}^{K} \pi_k \boldsymbol{N}(\boldsymbol{x}|\boldsymbol{\mu}_k, \boldsymbol{\Sigma}_k) \quad \text{Equation 1}$$

$$p(\boldsymbol{X}) = \prod_{n=1}^{N} P(\boldsymbol{x}_n) = \prod_{n=1}^{N} \sum_{k=1}^{K} \pi_k \boldsymbol{N}(\boldsymbol{x}|\boldsymbol{\mu}_k, \boldsymbol{\Sigma}_k) \quad \text{Equation 2}$$

Assuming $N$ trajectory points are collected, the joint probability of all observations is shown in Equation 2**.** The Expectation-Maximization (EM) algorithm is typically applied to find the optimal parameters with the maximum likelihood. The E-step computes the expectation of *Chen, Luo and Feng*the log-likelihood function with currently estimated parameters, and M-step tries to estimate new parameters that maximize the expected log-likelihood function in the E-step. The iterative process converges when the likelihood function's value differences are smaller than a predefined threshold.

The fitted GMM model can be used to 1) classification: assign each observation (i.e., trajectory point) to one of the components (or clusters) based on the means and covariance matrixes; and 2) prediction: generates (samples) new data points that follow the distribution, which can be used to determine the boundaries of the drivable area

**Occupancy grid map generation**

In mapping drivable areas, we need a proper representation of the road area for the planner to generate reference trajectories (*28*). A OGM is a mapping area representation that indicates which areas in the map are drivable or not. The mapping area is generally divided into grids with a pre-determined resolution. For the naïve binary occupancy grid map represented by a 2D matrix, each cell can either be occupied (1) or empty (0). For autonomous driving, the drivable area is defined as the area not occupied by the detection results (e.g., point cloud from LiDAR). For SLAM based mapping method, if an object is detected at the (x, y) location, then the cell containing this location is marked as non-drivable.

In our case, the locations covered by the GMM-generated trajectory points are drivable areas. As a result, the grids that are occupied by the trajectory points are marked as empty (0) while other uncovered grids are marked as occupied (1), which we call it "inverse occupancy grid map."

**Vehicle Trajectory Generation and Following**

With the OGM, CAV can generate a reference trajectory. We apply the Hybrid A* algorithm (*26*) (*27*) embedded in the Autoware.AI to generate the reference trajectory considering the vehicle dynamics and size by assigning a minimum turning radius.

We implement a longitudinal constant speed controller on the vehicle using PID control, which takes the difference between the reference speed and current speed and calculates the expected controller input based on the proportion of the error, the derivative of the error, and the integral of the error for ten most recent timesteps. The control command for the vehicle is then converted to the throttle and break command. Lateral speed control is based on steering and is calculated using the pure pursuit algorithm (*29*).

As shown in Figure 3, the pure pursuit algorithm simplifies a four-wheel vehicle into a two-wheel bicycle. The rear axle center is used as the reference point of the target vehicle for calculating lateral displacement. Then, it utilizes a geometric relationship between the rear wheel and target points to generate steering angles.

In our implementation, we send throttle, brake, and steering wheel control commands to the ego vehicle, which is exactly how a vehicle-level middleware handles the control task. We determine the target look ahead point at each planning step by both the reference trajectory and a predefined minimum lookahead distance $l_{dmin}$. The real lookahead distance is then calculated as $l_d$, the geometric distance between the lookahead reference point and the vehicle's current location. We calculate the steering angle $\delta$, using angle $\alpha$ between the vehicle's heading direction and lookahead direction, lookahead distance $l_d$, vehicle length $L$ and path curvature K by applying Equation 6:

$$\delta = \tan^{-1}\left(\frac{2L \sin \alpha}{l_d}\right) \qquad \text{Equation 3}$$

where

$$k = \frac{2 \sin \alpha}{l_d} \quad \delta = \tan^{-1} kL$$

**CASE STUDY**
To evaluate the feasibility of the proposed framework, we construct a work zone in the CARLA simulation platform. We established a work zone in the CARLA simulation platform to justify our argument that information from crowdsourcing trajectories can be integrated into the mapping process. This work zone scenario is constructed following the Manual on Uniform Traffic Control Devices (MUTCD), which will be discussed later. This scenario is used to our proposed mapping pipeline and compares it with the standard practice of creating HD maps from SLAM.

**Crowdsourcing Trajectory Based Map Generation**
The work zone is set up according to the MUTCD. Speed limit and width of offset are used to calculate the taper length (*30*). The speed limit is set to 60 mph. with merging taper 720 feet, shifting taper 360 feet, shoulder taper 240 feet, with activity area of 650 feet 36 feet in longitudinal and lateral directions. The setup of work zone is illustrated in figure 4.

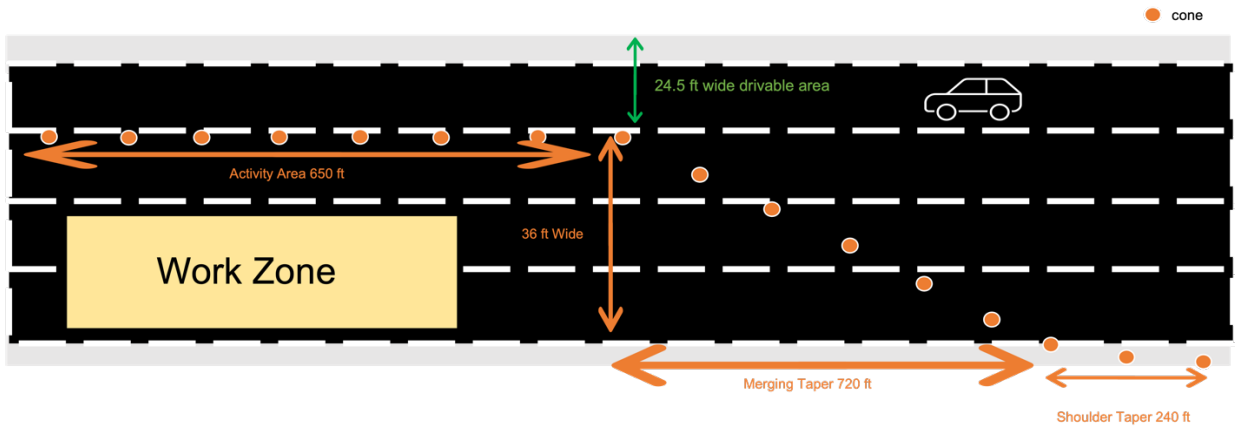

**Figure 2 an Illustrative figure of the workzone setup in the case study**

After the work zone was set up, we recruited a few participants to drive a vehicle in the CARLA simulation through the work zone using the keyboard. A total number of 30 trajectories are collected.

*Chen, Luo and Feng*

The collected trajectories are fed into a GMM model, and the number of total components (clusters) *K* are determined by the lowest AIC and BIC calculated from the fitted model. We tested different values of *K*, as shown in Figure 5, and the best value of *K* is equal to 15.

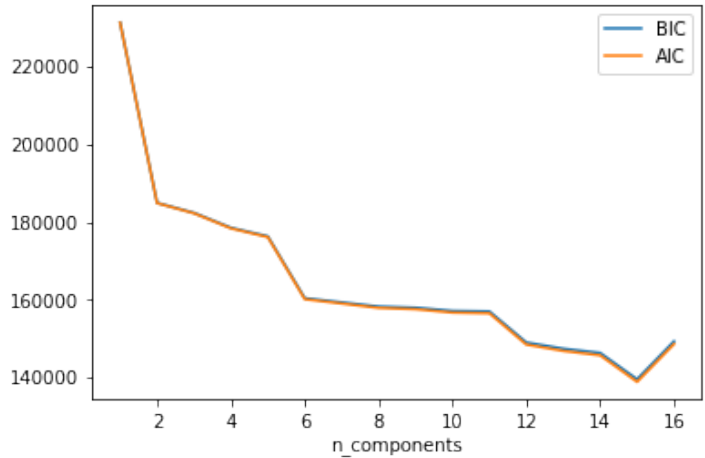

**Figure 3 Tests on different components of GMM**

Figure 5 shows the GMM clustering results with 15 components based on five crowdsourcing trajectories. More analysis on number of trajectories needed to build the GMM will be provided later.

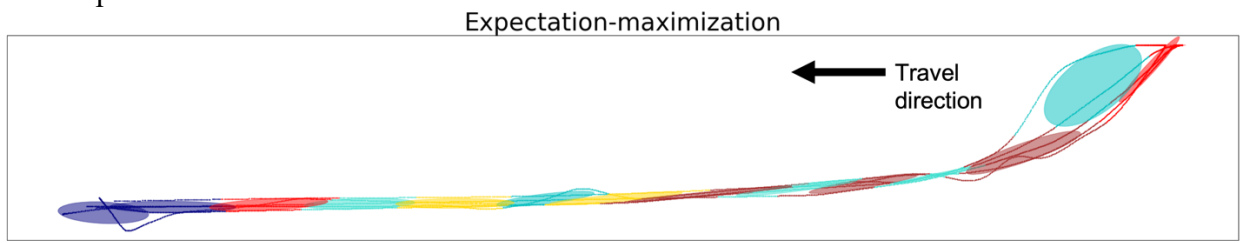

**Figure 4 GMM clustering result based on five collected trajectories**

After the GMM is obtained, 5000 points are sampled from the distribution within the 95% confidence interval to represent the drivable area, as shown in **Figure 6**.

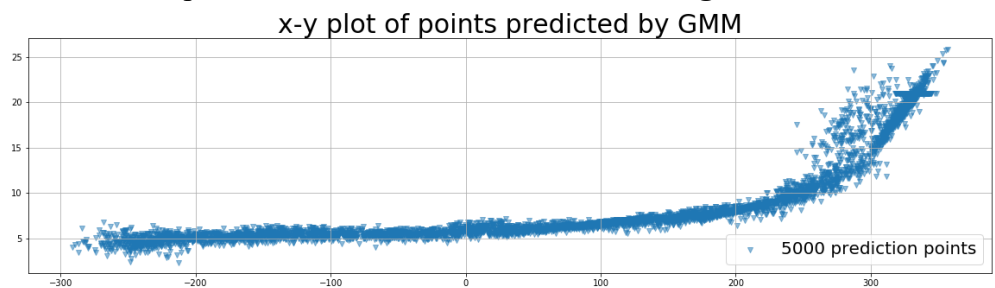

**Figure 5 Sample points generated from the GMM**

The OGM is produced based on the sample points shown in Figure 6, which are the prediction results from the GMM obtained. Note that we adjusted the size of each sample point to 5 by 5 inch in the plot. Finally, we process the IOGM from the inversed image shown in Figure 7. The adjusted sample points are converted to the OGM where white represents the drivable area.

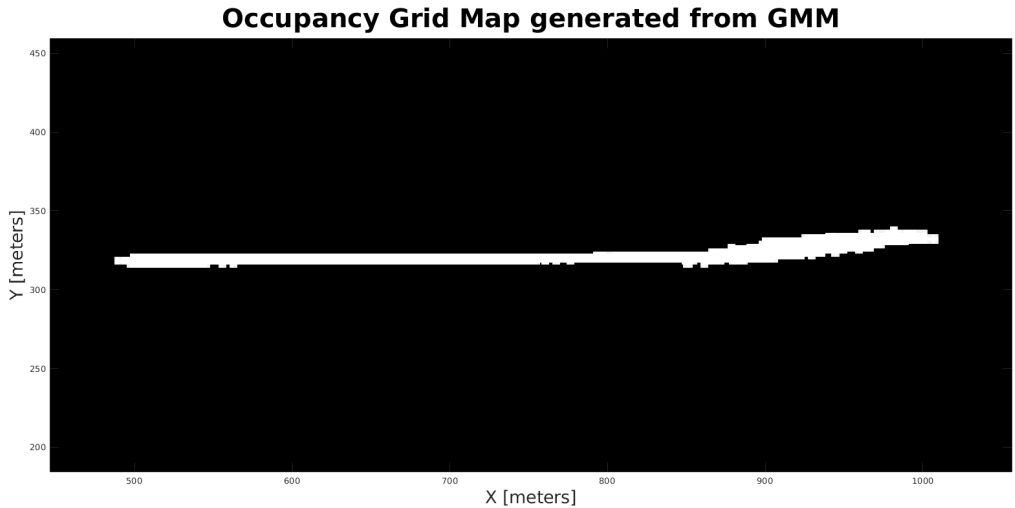

**Figure 6 OGM generated from crowdsourcing trajectories.**

**Benchmark AV Perception Model**

The benchmark for comparison follows the standard mapping pipeline for autonomous vehicles. We build the OGM of the work zone using the SLAM module provided by MATLAB (*31*)To collect the perception data for mapping, including LiDAR, Inertial Measurement Unit (IMU), and position data from the Global navigation satellite system (GNSS), we manually drive the ego vehicle through the work zone in the CARLA simulation platform.

To ensure the quality of the work zone detection, we perform ground removal in our pipeline to ensure the quality of mapping results from the Simple Morphological Filter SMRF algorithm (*25*). SMRF separates the ground and non-ground areas using structured elements. Given the elevation threshold, it first calculates a minimum elevation surface map, then calculates the slope between each area in the open and elevation surface maps. The area with a difference greater than the elevation threshold is classified as a non-ground area. In the SMRF algorithm, we set the parameter Elevation Threshold to 0.2. This is to ensure that point cloud data on traffic cones are not removed and best describe the temporarily changed drivable area. After executing ground removal, we align successive lidar point cloud data using feature-based registration.

Then we perform LiDAR Odometry and Mapping using the processed point cloud data along with the collected GPS and IMU data (*8*), which does not require the building of a pose graph, and it is closer to our application when a CAV is driving in a work zone for mapping. This way, the lidar point cloud is accumulated based on the trajectory collected from the ego vehicle. We then obtain the accumulated point cloud representing the work zone area. The occupancy grid map is obtained from the bird-eye-view projection of the accumulated point cloud, as shown in Figure 8.

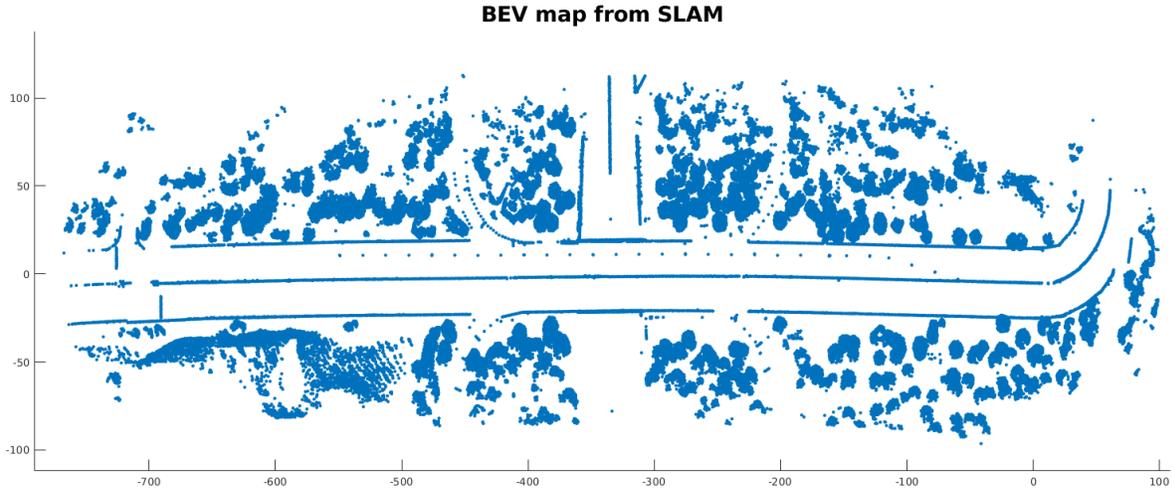

**Figure 7 OGM generated from Lidar point cloud by SLAM**

The algorithm for obtaining OGM from the point cloud is shown below:
**Algorithm: SLAM-based mapping from LiDAR and localization data**

**Input**: point cloud data from LiDAR with all time-stamps $\boldsymbol{P}_t^o$, with sensor locations $\boldsymbol{L}_t = \{x_t, y_t, z_t\}$, and $t \in \boldsymbol{T}$. Denote the first point cloud obtained as $P_0^o$, related initial sensor location $\boldsymbol{L_0} = \{x_0, y_0, z_0\}$;
**Begin**:
**For** *t* in $\boldsymbol{T}$ **do**
 Obtain point cloud $P_t^o$, location $L_t$
 Obtain Ground removed point cloud $P_t^p$ from $P_t^o$ using SMRF (*25*)
 Obtain transformation $T_t$ from $P_{t-1}^p$ and $P_t^p$ using LOAM point cloud registration(*8*)
 Update the map using $\boldsymbol{P_t^p}, \boldsymbol{T_t}$, and $\boldsymbol{T_{t-1}}$, and get the transformed point cloud $P_t^t$
 Align and merge the transformed point cloud $P_t^t$ into accumulated point cloud map with location $L_t$
Generate OGM from the bird-eye view of the accumulated point cloud data
**End**

 Where $\boldsymbol{P}_t^o$ is the unprocessed point cloud data collected at time *t*; $\boldsymbol{P}_t^p$ is the processed point cloud data at time *t*; and $T_t$ is the transformation to transfer point cloud from $P_t^p$ to $P_{t-1}^p$, making sure they are properly aligned. After updating the map, the newly obtained point cloud $P_t^p$ is transformed to the ego view of the first point cloud frame, which is denoted as $P_t^p$. Then the transformed point cloud at time t is accumulated to the point cloud map built from time 0 to time t.

 For the standard practice of creating an HD map using lidar (referred to as "benchmark" in the paper), we set up an ego vehicle equipped with 128-layer LiDAR, high-precision GNSS, and IMU for localization. We manually drive the test vehicle through the work zone

 For both proposed and standard mapping pipeline, we inflate the obtained OGM by 0.7 meters to ensure there is no collision between the side of the vehicle and the road boundary when generating the reference trajectories. For the implementation on PID controller for longitudinal speed control, we used KP = 0.5, KI = 0.018 and KD = 0.4, also we set vehicle speed $v = 10 m/s$ as the constant overall speed. For pure pursuit on calculating vehicle turning angle at each timestep, we used $l_{dmin} = 3\ meters$, vehicle length $L = 4.6391\ meters$, and a constant speed for vehicle $v = 10 m/s$ to calculate the vehicle turning angle.

**RESULTS AND COMPARISON**
We evaluate the proposed method from different perspectives. First, we perform a pixel-level evaluation to show how well our method-generated drivable area overlaps with the real drivable area. This evaluation demonstrates that information provided by crowdsourcing trajectory can assist drivable area inference for CAVs.

Figure 9 shows a pixel-wise comparison with the SLAM map. The blue dots represent the BEV map obtained from SLAM. The red area refers to the drivable area inference from the crowdsourcing trajectories. The black line is the planned reference trajectory using the drivable area inferenced from the crowdsourcing trajectories. The area in the dashed green lines shows the real drivable area. By comparing the obtained drivable area with the ground truth drivable area, we can get a precision of 90.08% for the drivable area we infer with 5 crowdsourcing trajectories. The precision is defined by as the ratio between the true positive (TP) areas and the summation of TP areas and false positive (FP) areas. TP areas are pixels (cells) correctly defined as the drivable areas, while FP areas refer to the areas we thought is drivable but actually not. Based on the identified drivable areas, the path planner generated a feasible reference trajectory that 1) follow the traffic rules; and 2) keep safety distances between the traffic cones on the left and the road boundary on the right.

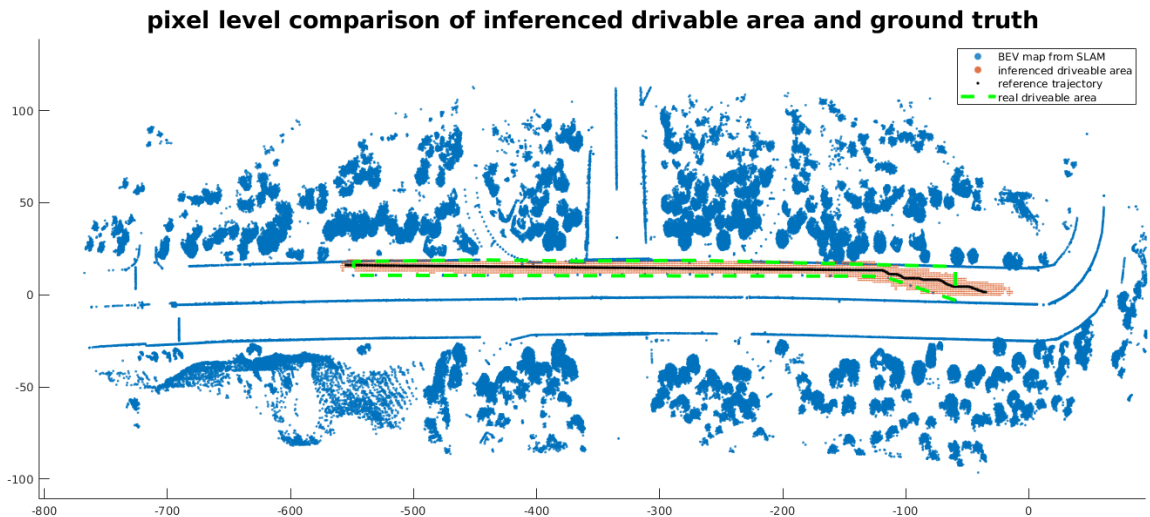

**Figure 8 Pixel-wise comparison with the SLAM data for the drivable area.**

To compare with the current standard single vehicle solution, we also show the reference trajectory points generated without crowdsourcing information, by only using the SLAM-based map, as shown in Figure 10. It clearly showed that the path planner could not interpret the meaning of the traffic cone and violates safety rules by driving directly through the cones.

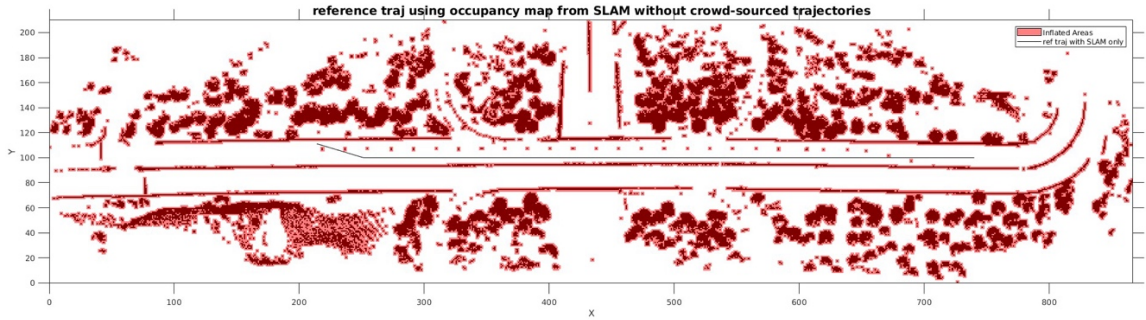

**Figure 9 Reference Trajectory Generation Without Information From the Crowdsourcing Trajectories.**

Second, the proposed method is evaluated from the safety's perspective which is the most critical criteria in CAV navigation. . Minimum distances of the CAV to the work zone and road boundary are used as the performance metric. This minimum distances are measured using the reference trajectory.

Figure 11 shows the location of the CAV when driving through the work zone. The orange line represents the center of the vehicle, while the green and red lines represent the left and right boundary of the vehicle respectively. The blue dots are the location of the traffic cones and the purple line is the roadway boundary. The minimum distance of the from the left boundary of the vehicle to the cone is 1.297 meters. The minimum distance from the right boundary of the CAV to the road boundary including shoulder is 3.50 meters. This evaluation shows that our method ensures the safety of CAV by keeping reasonable distances from both sides through of the work zone. The plot also shows that the distances between the vehicle and the cones/road boundary do not fluctuate a lot, which shows a good stability of the path planning model.

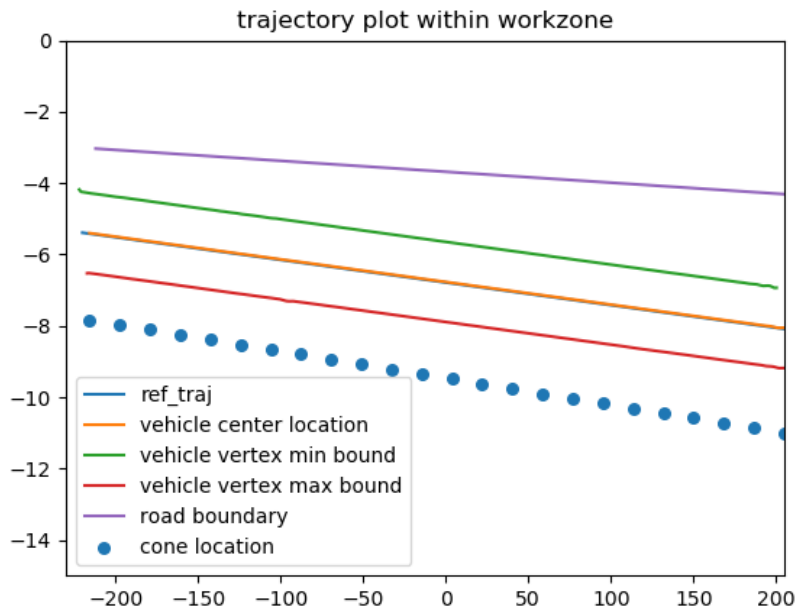

*Chen, Luo and Feng*

**Figure 10 Trajectory plot of CAV within the work zone**

**Discussion and future work**

In this paper, we demonstrate the concept of utilizing crowdsourcing vehicle trajectories to improve the mapping of CAVs. This preliminary proof of concept can be extended in several directions.

The number of crowdsourcing trajectories needed to accurately estimate the geometry of a drivable area is a key question. The more trajectories used to build the GMM model, the more accurate the estimated OGM will be. However, more trajectories require more time to collect, given the penetration rate of crowdsourcing vehicles that pass the work zone. The optimal number of required trajectories needs to be investigated, which is closely related to the parameters in the GMM model (e.g., number of components), work zone geometry, and the required accuracy level. When the penetration rate of crowdsourcing vehicles or the total traffic volume is low, data augmentation may be needed to generate synthetic trajectory data.

The geometry of the work zones may change frequently. For example, a traffic cone is blown away by a truck, or the closed lane for construction is changed. If an HD map is used for navigation, then these maps need to be updated frequently (e.g., using a mapping vehicle), which introduces high costs. Another advantage of utilizing crowdsourcing trajectories is that cost is much lower. However, the same problem exists when using crowdsourcing trajectories to build a map with the frequently changed drivable area. A dynamic update mechanism is needed to determine whether the geometry is changed. For example, Bayesian optimization can be applied to determine potential changes and update the distribution based on new observations.

In our current case study, we didn't consider errors in GPS, which is one of the greatest concerns in crowdsourcing vehicle trajectories. According to an existing study, the GPS errors can reach more than 1 to 2 meters, which are collected from connected vehicles (*32*). This error level may already cause significant problems for CAV mapping and collision avoidance. Therefore, it is critical to incorporate uncertainties in the OGM generation to accommodate GPS errors. As a result, probabilistic OGM should be considered in the figure, where the occupation of a grid is represented by a probability, not just 1 or 0. Moreover, the probabilistic OGM should be integrated with the path planning module in generating reference trajectories.

**CONCLUSIONS**

In our work, we implemented a prototype of integrating crowdsourcing vehicle trajectories into the standard perception-planning pipeline of the autonomous vehicle in generating mapping information. We showed that information from crowdsourcing vehicle trajectories could improve CAV navigation in temporarily changed drivable areas. We used the work zone as an example. We showed that without additional information, current practice in autonomous driving has difficulties in safely navigating through a work zone without violating traffic rules. Our work demonstrated a promising new direction that utilizes the human driving experience to help autonomous vehicles perceive the complex driving environment.